\documentclass[sigconf]{acmart}

\usepackage{booktabs} 
\usepackage{threeparttable}

\usepackage{tabularx}
\usepackage{multirow}
\usepackage{subfigure}

\usepackage{amsmath}
\DeclareMathOperator*{\argmax}{arg\,max}

\copyrightyear{2019}
\acmYear{2019} 
\setcopyright{iw3c2w3}
\acmConference[WWW '19]{Proceedings of the 2019 World Wide Web Conference}{May 13--17, 2019}{San Francisco, CA, USA}
\acmBooktitle{Proceedings of the 2019 World Wide Web Conference (WWW'19), May 13--17, 2019, San Francisco, CA, USA}
\acmPrice{}
\acmDOI{10.1145/3308558.3313737}
\acmISBN{978-1-4503-6674-8/19/05}

\usepackage{balance}

\begin{document}
\title{Learning to Generate Questions by Learning\\What not to Generate} 

\author{Bang Liu$^1$, Mingjun Zhao$^1$, Di Niu$^1$, Kunfeng Lai$^2$, Yancheng He$^2$, Haojie Wei$^2$, Yu Xu$^2$}
       \affiliation{$^1$University of Alberta, Edmonton, AB, Canada}
 \affiliation{$^2$Platform and Content Group, Tencent, Shenzhen, China}

\begin{abstract}
Automatic question generation is an important technique that can improve the training of question answering, help chatbots to start or continue a conversation with humans, and provide assessment materials for educational purposes.
Existing neural question generation models are not sufficient mainly due to their inability to properly model the process of how each word in the question is selected, i.e., whether repeating the given passage or being generated from a vocabulary.
In this paper, we propose our Clue Guided Copy Network for Question Generation (CGC-QG), which 
is a sequence-to-sequence generative model with copying mechanism, yet employing a variety of novel components and techniques to boost the performance of question generation.
In CGC-QG, we design a multi-task labeling strategy to identify whether a question word should be copied from the input passage or be generated instead, guiding the model to learn the accurate boundaries between copying and generation. Furthermore, our input passage encoder takes as input, among a diverse range of other features, the prediction made by a clue word predictor, which helps identify whether each word in the input passage is a potential clue to be copied into the target question.   
The clue word predictor is designed based on a novel application of Graph Convolutional Networks onto a syntactic dependency tree representation of each passage, thus being able to predict clue words only based on their context in the passage and their relative positions to the answer in the tree.
We jointly train the clue prediction as well as question generation with multi-task learning and a number of practical strategies to reduce the complexity.
Extensive evaluations show that our model significantly improves the performance of question generation and out-performs all previous state-of-the-art neural question generation models by a substantial margin.
\end{abstract}



\maketitle

\section{Introduction}
\label{sec:intro}

Asking questions plays a vital role for both the growth of human beings and the improvement of artificial intelligent systems.
As a dual task of question answering, question generation based on a text passage and a given answer has attracted much attention in recent years.
One of the key applications of question generation is to automatically produce question-answer pairs to enhance machine reading comprehension systems~\cite{du2017learning,yuan2017machine,tang2017question,tang2018learning}. Another application is generating practice exercises and assessments for educational purposes \cite{heilman2010good, heilman2011automatic, danon2017syntactic}. Besides, question generation is also important in conversational systems and chatbots such as Siri, Cortana, Alexa and Google Assistant, helping them to kick-start and continue a conversation with human users \cite{mostafazadeh2016generating}.

\begin{figure}[tb]
\centering
\includegraphics[width=3.3in]{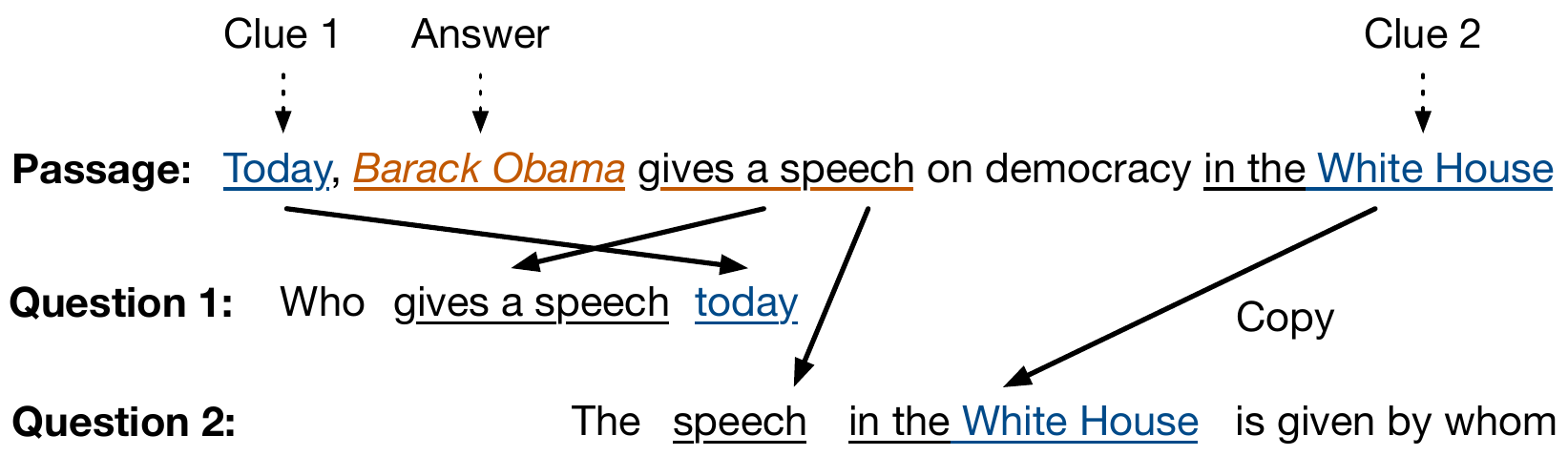}
\vspace{-4mm} 
\caption{Questions are often asked by repeating some text chunks in the input passage, while there is great flexibility as to which chunks are repeated.
}
\label{fig:CaseStudy}
\vspace{-4mm} 
\end{figure}

Conventional methods for question generation rely on heuristic rules to perform syntactic transformations of a sentence to factual questions  \cite{heilman2011automatic,chali2015towards}, following grammatical and lexical analysis. 
However, such methods require specifically crafted transformation and generation rules, with low generalizability. Recently, various neural network models have been proposed for question generation \cite{du2017learning,zhou2017neural,hu2018aspect,kim2018improving,gao2018difficulty}. These models formulate the question generation task as a sequence-to-sequence (Seq2Seq) neural learning problem, where different types of encoders and decoders have been designed.
Like many other text generation tasks, the copying or pointer mechanism \cite{gulcehre2016pointing} is also widely adopted in question generation to handle the copy phenomenon between input passages and output questions, e.g., \cite{zhou2017neural,yuan2017machine,subramanian2017neural,kim2018improving,song2018leveraging}.
However, we point out that a common limitation that hurdles  question generation models mainly use copying mechanisms to handle out-of-vocabulary (OOV) issues, and fail to mark a clear boundary between the set of question words that should be directly copied from the input text and those that should be generated instead.

In this paper, we generate questions by learning to identify where each word in the question should come from, i.e., whether generated from a vocabulary or copied from the input text, and given the answer and its context, what words can potentially be copied from the input passage.
People usually repeat text chunks in an input passage when asking questions about it, and generate remaining words from their own language to form a complete question.
For example, in Fig.~\ref{fig:CaseStudy}, given an input passage ``Today, Barack Obama gives a speech on democracy in the White House'' and an answer ``Barack Obama'', we can ask a question ``The speech in the White House is given by whom?'' Here the text chunks ``speech'' and ``in the White House'' are copied from the input passage.
However, in the situation that a vocabulary word is an overlap word between a passage and a question, existing models do not clearly identify the underlying reason about the overlapping. In other words, whether this word is copied from the input or generated from vocabulary is not properly labeled. For example, some approaches \cite{zhou2017neural} only take out-of-vocabulary (OOV) words that are shared by both the input passage and target question as copied words, and adopt a copying mechanism to learn to copy these OOV words into questions.
In our work, given a passage and a question in a training dataset, we label a word as copied word if it is a nonstop word shared by both the passage and the question, and its word frequency rank in the vocabulary is lower than a threshold.
We further aggressively shortlist the output vocabulary based on frequency analysis of the non-copied question words (according to our labeling criteria).
Combining this labeling strategy with target vocabulary reduction, during the process of question generation, our model can make better decisions about when to copy or generate, and predict what to generate more accurately.


After applying the above strategies, a remaining problem is which word we should choose to copy. For example, in Fig.~\ref{fig:CaseStudy}, we can either ask the question ``Who gives a speech today?'' or the question ``The speech in the White House is given by whom?'', where the two questions are related to two different copied text chunks ``today'' and ``White House''.
We can see that asking a question about a passage and a given answer is actually a ``one-to-many'' mapping problem: we can ask questions in different ways based on which subset of words we choose to copy from input.
Therefore, how to enable a neural model to learn what to copy is a critical problem. 
To solve this issue, we propose to predict the potential clue words in input passages.
A word is considered as a ``clue word'' if it is helpful to reduce the uncertainty of the model about how to ask a question or what to copy, such as ``White House'' in question 2 of Fig.~\ref{fig:CaseStudy}. In our model, we directly use our previously mentioned copy word labeling strategy to assign a binary label to each input word to indicate whether it is a clue word or not.

To predict the potential clue words in input, we have designed a novel clue prediction model that combines the syntactic dependency tree of an input with Graph Convolutional Networks. The intuition underlying our model design is that: the words copied from an input sentence to an output question are usually closely related to the answer chunk, and the patterns of the dependency paths between the copied words and answer chunks can be captured via Graph Convolutional Networks.
Combining the graphical representation of input sentences with GCN enables us to incorporate both word features, such as word vector representation, Named Entity (NE) tags, Part-of-Speech (POS) tags and so on, and the syntactic structure of sentences for clue words prediction. 

To generate questions given an answer chunk and the predicted clue words distribution in an input sentence, we apply the Seq2Seq framework which contains our feature-rich sentence encoder and a decoder with attention and copy mechanism. The clue word prediction results are incorporated into the encoder by a binary feature embedding.
Based on our multi-task labeling strategy, i.e., labeling for both clue prediction and question generation, we jointly learn different components of our model in a supervised manner.

We performed extensive evaluation on two large question answering datasets: the SQuAD dataset v1.1 \cite{rajpurkar2016squad} and the NewsQA dataset \cite{trischler2016newsqa}. For each dataset, we use the answer chunk and the sentence that contains the answer chunk as input, and try to predict the question as output.
We compared our model with a variety of existing rule-based and neural network based models. Our approach achieves significant improvement by jointly learning the potential clue words distribution for copy and the encoder-decoder framework for generation, and out-performs state-of-the-art approaches.


\begin{figure}[tb]
\centering
\includegraphics[width=3.4in]{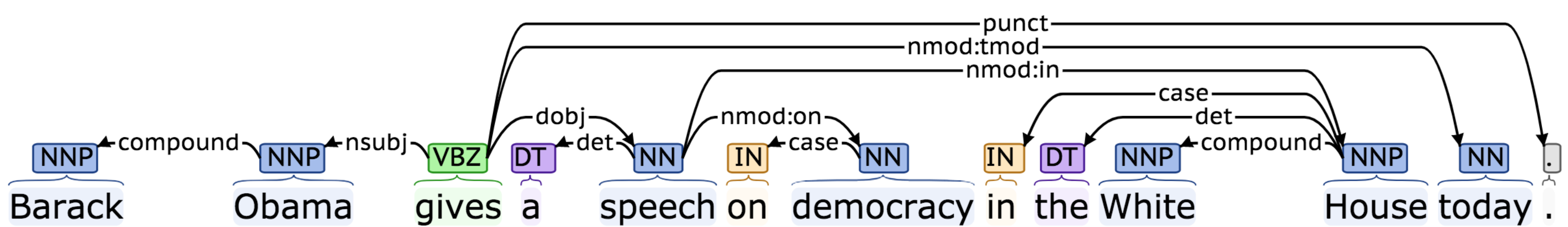}
\vspace{-4mm} 
\caption{An example to show the syntactic structure of an input sentence. Clue words ``White House'' and ``today'' are close to the answer chunk ``Barack Obama'' with respect to the graph distance, though they are not close to each other in terms of the word order distance.}
\label{fig:syntax}
\vspace{-4mm} 
\end{figure}
 
\section{Problem Definition and Motivation}
\label{sec:data}

In this section, we formally introduce the problem of question generation, and illustrate the motivation behind our work.

\subsection{Answer-aware Question Generation}
Let us denote a passage by $P$, a question related to this passage by $Q$, and the answer of that question by $A$. A passage can be either an input sentence or a paragraph, based on different datasets. A passage consists of a sequence of words $P = \{p_t\}_{t=1}^{|P|}$ where $|P|$ denotes the length of $P$. A question $Q = \{q_t\}_{t=1}^{|Q|}$ contains words from either a predefined vocabulary $V$ or from the input text $P$. The task is finding the most probable question $\hat{Q}$ given an input passage and an answer:

\begin{equation}
\label{eq:problem}
\hat{Q} = \argmax_Q prob(Q|P, A).
\end{equation}

\begin{figure}[tb]
\centering
\includegraphics[width=2.9in]{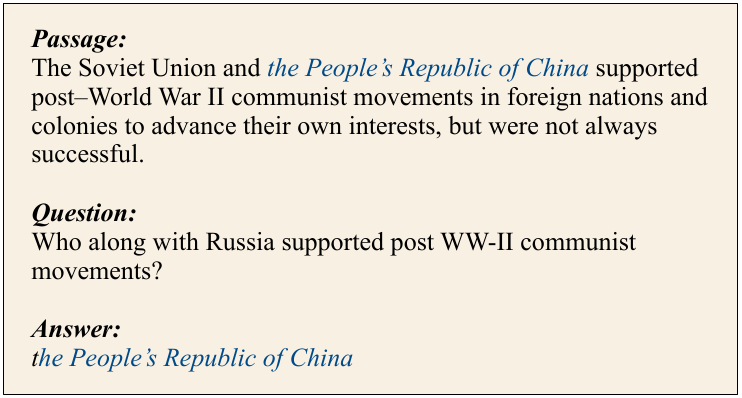}
\vspace{-3mm} 
\caption{An example from the SQuAD dataset. Our task is to generate questions given an input passage and an answer. In SQuAD dataset, answers are sub spans of the passages.}
\label{fig:example}
\vspace{-4mm} 
\end{figure}

Fig.~\ref{fig:example} shows an example of the dataset used in our paper. Note that the answer $A$ of the question $Q$ is limited to sub spans of the input passage. However, our work can be easily adapted to the cases where the answer is not a sub span of the input passage by adding an extra answer encoder.

\subsection{What to Ask: Clue Word Prediction}
\label{subsec:what-to-ask}
Even given the answer of a desired question, the task of question generation is still not a one-to-one mapping, as there are potentially multiple aspects related to the given answer. For example, in Fig.~\ref{fig:syntax}, given the answer chunk ``Barack Obama'', the questions ``The speech in the White House is given by whom?'', ``Who gives a speech on democracy?'', and ``Today who gives a speech?'' are all valid questions. As we can see, although these questions share the same answer, they can be asked in different ways based on which word or phrase we choose to copy (e.g., ``White House'', ``democracy'', or ``today'').

To resolve this issue, we propose to predict the potential ``clue words'' in the input passage. A ``clue word'' is defined as a word or phrase that is helpful to guide the way we ask a question, such as ``White House'', ``democracy'', and ``today'' in Fig.~\ref{fig:syntax}. We design a clue prediction module based on the syntactical dependency tree representation of passages and Graph Convolutional Networks. Graph Convolutional Networks generalize Convolutional Neural Networks to graph-structured data, and have been successfully applied to natural language processing tasks, including semantic role labeling \cite{marcheggiani2017encoding}, document matching \cite{liu2018matching,zhang2018multiresolution}, relation extraction \cite{zhang2018graph}, and so on.
The intuition underlying our model design is that clue words will be more closely connected to the answer in dependency trees, and the syntactical connection patterns can be learned via graph convolutional networks. For example, in Fig.~\ref{fig:syntax}, the graph path distances between answer ``Barack Obama'' to ``democracy'', ``White House'', and ``today'' are $3$, $3$, and $2$, while the corresponding word order distances between them are $5$, $8$, and $10$, respectively.

\subsection{How to Ask: Copy or Generate}
Another important problem of question generation is when to choose a word from the target vocabulary and when to copy a word from the input passage during the generation process. 
People tend to repeat or paraphrase some text pieces in a given passage when they ask a question about it. For instance, in Fig.~\ref{fig:example}, the question ``Who along with Russia supported post WW-II communist movements'', the text pieces ``supported'', ``post'', and ``communist movements'' are repeating the input passage, while ``Russia'' and ``WW-II'' are synonymous replacements of input phrases ``The Soviet Union'' and ``World War II''. Therefore, the copied words in questions should not be restricted to out-of-vocabulary words.
Although this phenomenon is well-known by existing approaches, they do not properly and explicitly distinguish whether a word is from copy or from generate.

In our model,
we consider the non-stop words that are shared by both an input passage and the output question as clue words, and encourage the model to copy clue words from input.
After labeling clue words, we train a GCN-based clue predictor to learn whether each word in source text can be copied to the target question. The predicted clue words are further fed into a Seq2Seq model with attention and copy mechanism to help with question generation.
Besides, different from existing approaches that share a vocabulary between source passages and target questions, we reduce the size of the target vocabulary of questions to be smaller than source passages and only include words with word frequency higher than a threshold. The idea is that the low-frequency words in questions are usually copied from source text rather than generated from a vocabulary.
By combining the above strategies, our model is able to learn when to generate or copy a word, as well as which words to copy or generate during the progress of question generation.
We will describe our operations in more detail in the next section.

\section{Model Description}
\label{sec:model}

\begin{figure*}[tb]
\centering
\includegraphics[width=7.0in]{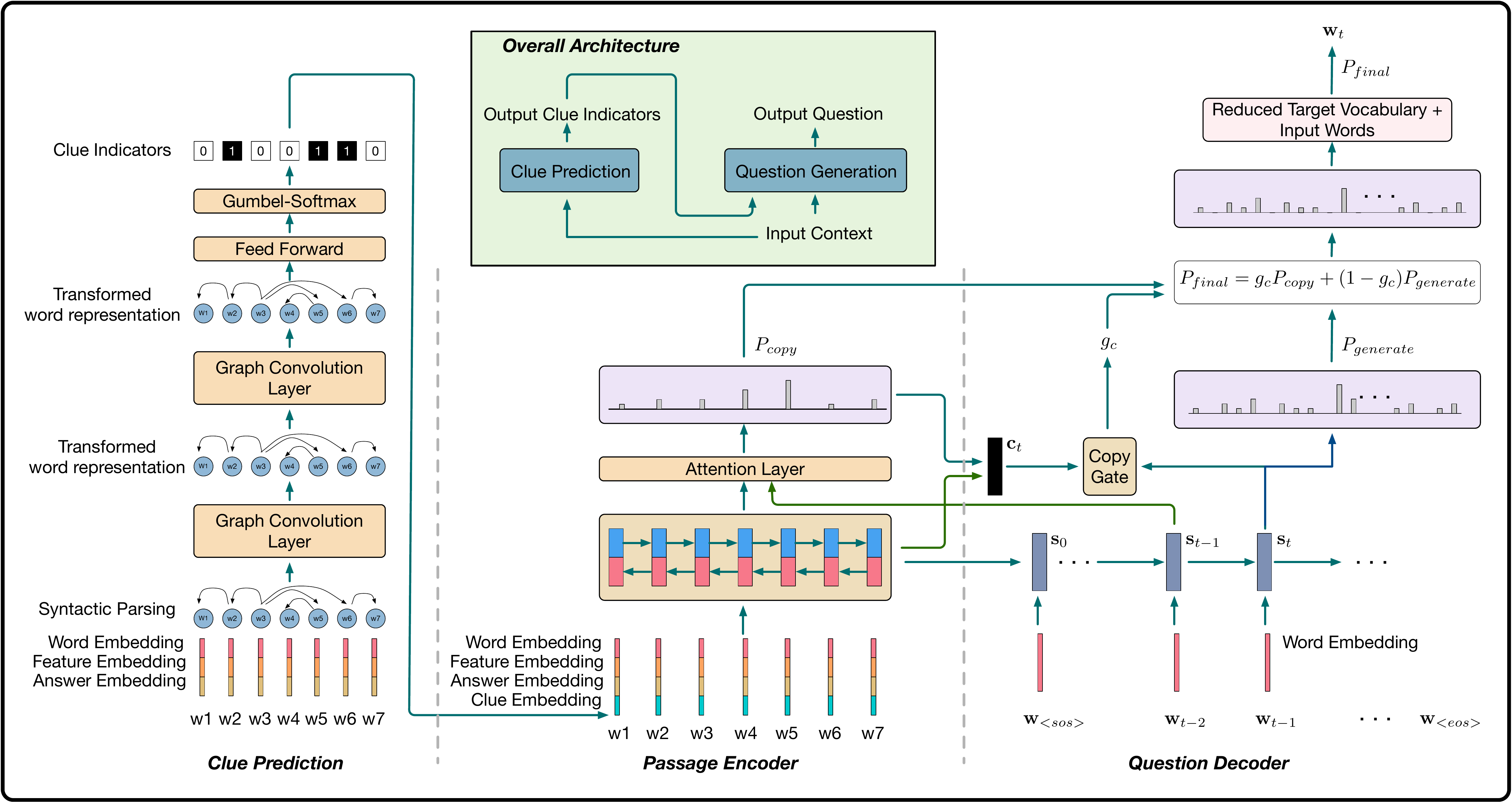}
\vspace{-4mm} 
\caption{Illustration of the overall architecture of our proposed model. It contains a GCN-based clue word predictor, a masked feature-rich encoder for input passages, and an attention-and-copy-based decoder for generating questions.}
\label{fig:system}
\vspace{-4mm} 
\end{figure*}

In this section, we introduce our proposed framework in detail. Similar to \cite {zhou2017neural,serban2016generating,du2017learning}, our question generator is based on an encoder-decoder framework, with attention and copying mechanisms incorporated. Fig.~\ref{fig:system} illustrates the overall architecture and detailed components in our model. Our model consists of three components: the clue word predictor, passage encoder and question decoder. 

The clue word predictor predicts potential clue words that may appear a target question, based on the specific context of the input passage (without knowing the target question). It utilizes a syntactic dependency tree to reveal the relationship of answer tokens relative to other tokens in a sentence. Based on the tree representation of the input passage, we predict the distribution of clue words by a GCN-based encoder applied on the tree and a Straight-Through (ST) Gumbel-Softmax estimator \cite{jang2016categorical} for clue word sampling.
The outputs of the clue word predictor are fed to the passage encoder to advise the encoder about what words may potentially be copied to the target question.

The passage encoder incorporates both the predicted clue word distribution and a variety of other feature embeddings of the input words, such as lexical features, answer position indicators, etc. Combined with a proposed low-frequency masking strategy (a shortlist strategy to reduce complexity on tuning input word embeddings), our encoder learns to better capture the useful input information with fewer trainable parameters.

Finally, the decoder jointly learns the probabilities of generating a word from vocabulary and copying a word from the input passage. At the decoder side, we introduce a multitask learning strategy to intentionally encourage the copying behavior in question generation. This is achieved by explicitly labeling the copy gate with a binary variable when a (non-stop) word in the question also appears in the input passage in the training set. Other multi-task labels are also incorporated to accurately learn when and what to copy.    
We further shortlist the target vocabulary based on the frequency distribution of non-overlap words. These strategies, assisted by the encoded features, help our model to clearly learn which word in the passage is indeed a clue word to be copied into the target question, while generating other non-clue words accurately.

The entire model is trained end-to-end via multitask learning, i.e., by minimizing a weighted sum of losses associated with different labels. In the following, we first introduce the encoder and decoder, followed by a description of the clue word predictor.




\subsection{The Passage Encoder with Masks}
Our encoder is based on bidirectional Gated Recurrent Unit (BiGRU) \cite{chung2014empirical}, taking word embeddings, answer position indicators, lexical and frequency features of words, as well as the output of the clue word predictor as the input. Specifically, for each word $p_i$ in input passage $P$, we concatenate the following features to form
a concatenated representation $w_i$ to be input into the encoder:
\begin{itemize}
	\item \textbf{Word Vector}. We initialize each word vector by Glove embedding \cite{pennington2014glove}. If a word is not covered by Glove, we initialize its word vector randomly.
	\item \textbf{Lexical Features}. We perform Named Entity Recognition (NER), Part-of-Speech (POS) tagging and Dependency Parsing (DEP) on input passages using spaCy \cite{spacy}, and concatenate the lexical feature embedding vectors.
	\item \textbf{Binary Features}. We check whether each word is lowercase or not, whether it is a digit or like a number (such as word ``three''), and embed these features by vectors.
	\item \textbf{Answer Position}. Similar to \cite{zhou2017neural}, we utilize the B/I/O tagging scheme to label the position of a given answer, where a word at the beginning of an answer is marked with \textbf{B}, \textbf{I} denotes the continuation of the answer, while words not contained in an answer are marked with \textbf{O}.
	\item \textbf{Word Frequency Feature}. We derive the word vocabulary from passages and questions in the training dataset. We then rank all the words in a descending order in terms of word frequencies, such that the first word is the most frequent word. The top $r_h$ words are labeled as frequent words. Words ranked between $r_h$ and $r_l$ are labeled as intermediate words. The remaining with rank lower than $r_l$ are labeled as rare words, where $r_h$ and $r_l$ are two predefined thresholds. In this way, each word will be assigned a frequency tag \textbf{L} (low frequency), \textbf{H} (highly frequent) or \textbf{M} (medium frequency).
	\item \textbf{Clue Indicator Feature}. In our model, the clue predictor (which we will introduce in more detail in Sec.~\ref{subsec:cluepredict}) assigns a binary value to each word to indicate whether it is a potential clue word or not.
\end{itemize}

Denote an input passage by $P = (w_1, w_2, \cdots, w_{|P|})$. The BiGRU encoder reads the input sequence $w_1, w_2, \cdots, w_{|P|}$ and produces a sequence of hidden states $h_1, h_2, \cdots, h_{|P|}$ to represent the passage $P$. Each hidden state is a concatenation of a forward representation and a backward representation:
\begin{align}
\label{eq:bigru}
h_i &= [\overrightarrow{h}_i; \overleftarrow{h}_i],\\
\overrightarrow{h}_i &= \mathbf{BiGRU}(w_i, \overrightarrow{h}_{i-1}),\\
\overleftarrow{h}_i &= \mathbf{BiGRU}(w_i, \overleftarrow{h}_{i+1}),
\end{align}
where  $\overrightarrow{h}_i$ and $\overleftarrow{h}_i$ are the forward and backward hidden states of the $i$-th token in $P$, respectively.
 
Furthermore, rather than learning the full representation $w_i$ for every word in the vocabulary, we use a masking strategy to replace the word embeddings of low-frequency words with a special <l> token, such that the information of low-frequency words is only represented by its answer/clue indicators and other augmented feature embeddings, except the word vectors.
This strategy can improve performance due to two reasons.
First, the augmented tagging features of a low-frequency word tend to be more influential than the word meaning in question generation. 
For example, given a sentence ``<PERSON> likes playing football.'', a question that can be generated is ``What does <PERSON> like to play?''---what the token ``<PERSON>'' exactly is does not matter.  This way, the masking strategy helps the model to omit the fine details that are not necessary for question generation. Second, the number of parameters that need be learned, especially the number of word embeddings that need be tuned, is largely reduced. Therefore, masking out unnecessary word embeddings while only keeping the corresponding augmented features and indicators does not hurt the performance of question generation. It actually improves training by reducing the model complexity.

\begin{figure*}[t]
                        \centering
                        \subfigure[Rank Distribution of All Question Words]{
                \includegraphics[width=2.2in]{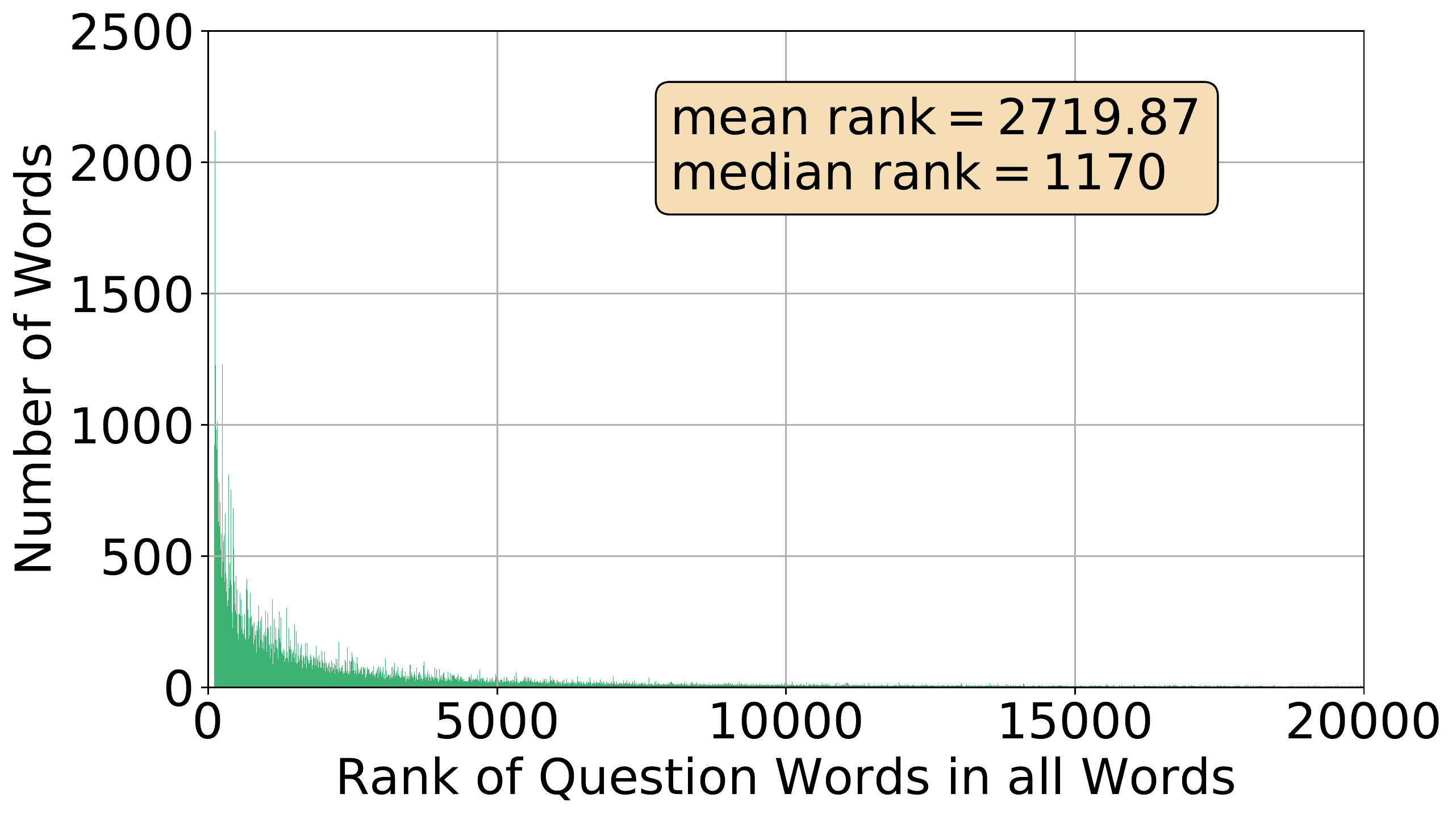}
                                \label{fig:all_word_ranks}
                        }
                \hspace{-1mm}
                        \subfigure[Rank Distribution of Generated Question Words]{
                \includegraphics[width=2.2in]{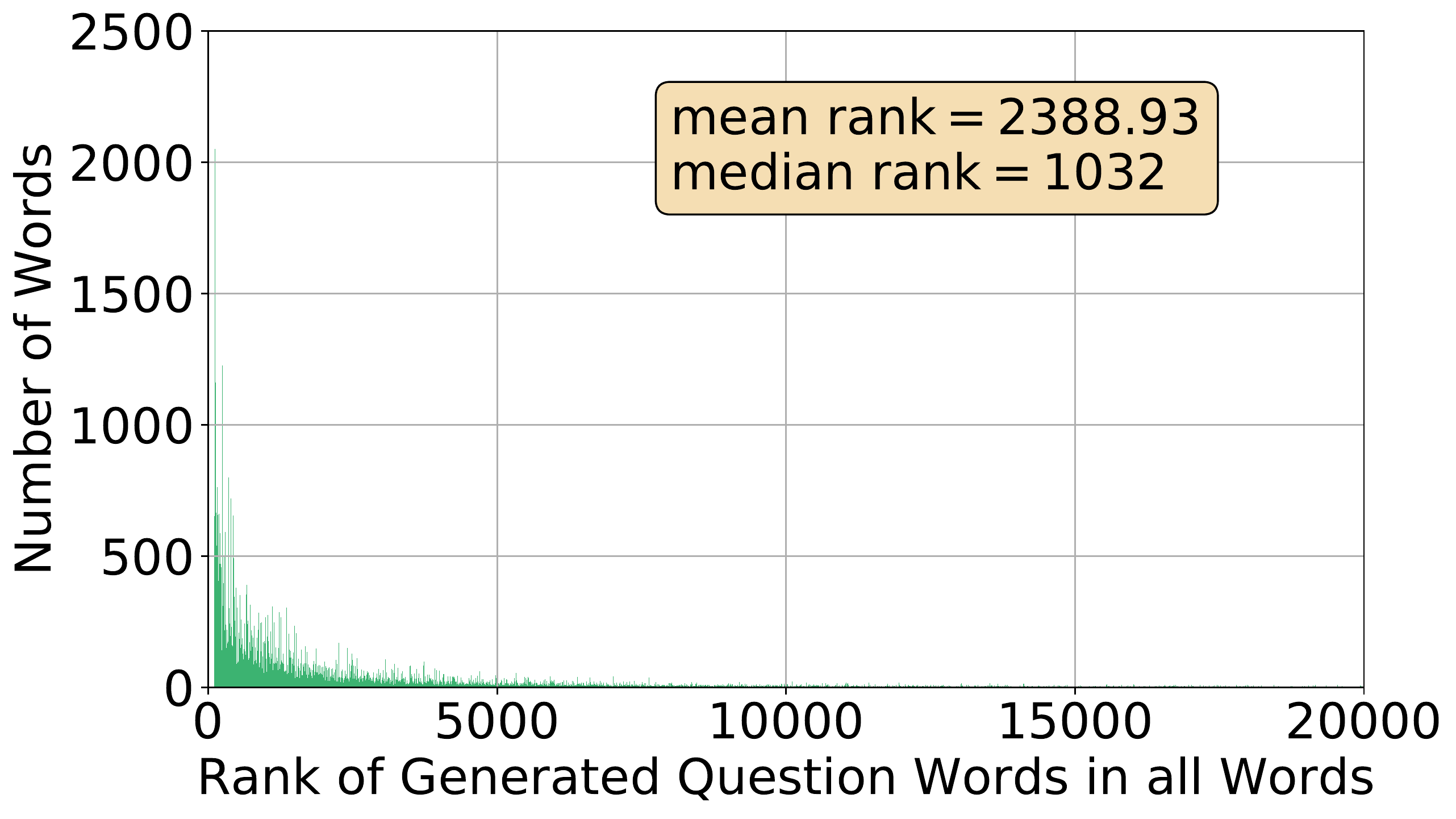}
                                \label{fig:not_overlap_ranks}
                        }
                        \hspace{-1mm}
                        \subfigure[Rank Distribution of Copied Question Words]{
                \includegraphics[width=2.2in]{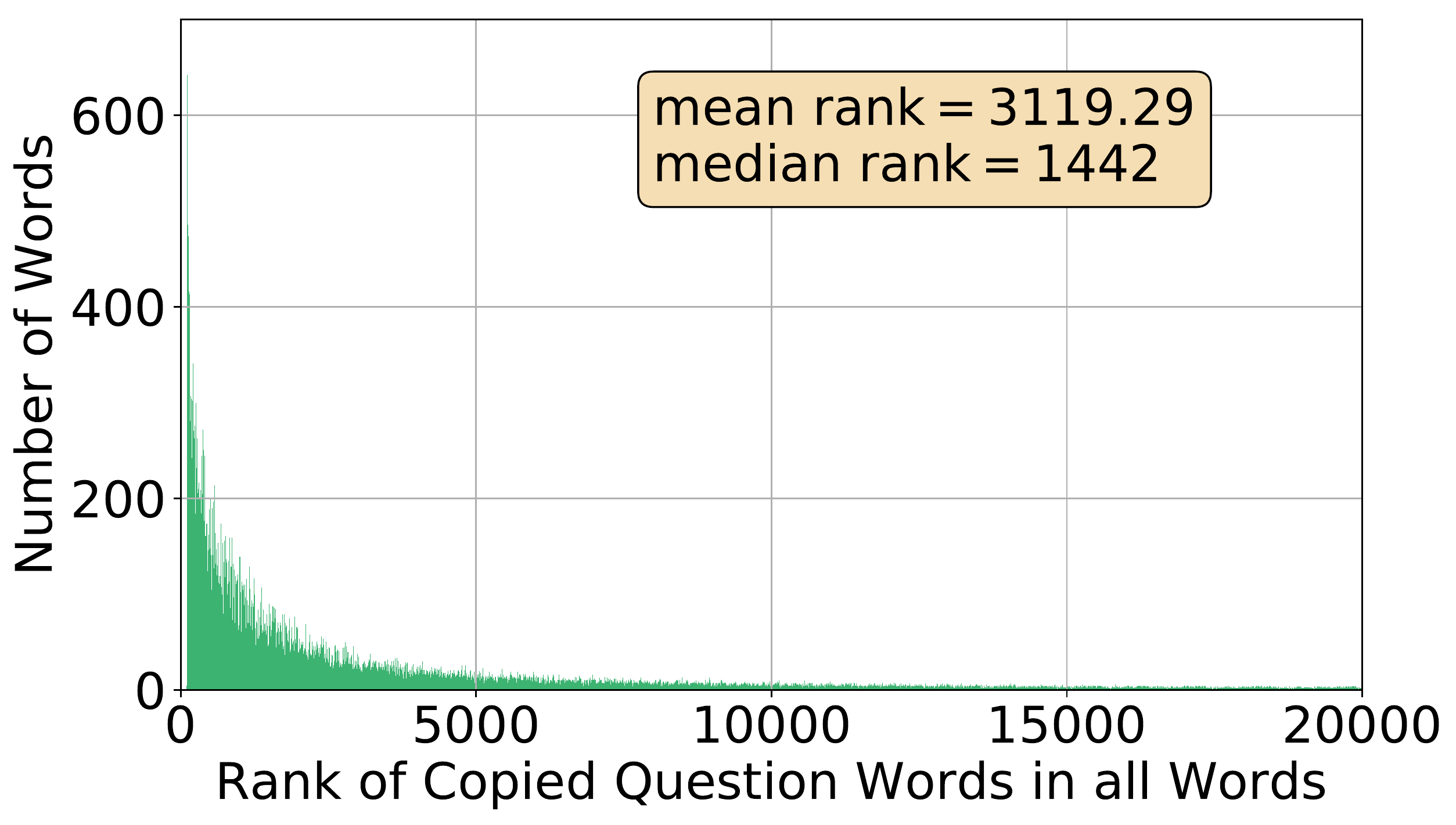}
                                \label{fig:non_stopword_overlap_ranks}
                        }
                \vspace{-4mm}
                \caption{Comparing the rank distributions of all question words, words from generation, and words from copy.}
                \label{fig:rank-hist}
\vspace{-4mm}
\end{figure*}

\subsection{The Question Decoder with Aggressive Copying}

In the decoding stage, we utilize another GRU with copying mechanism to generate question words sequentially based on the encoded input passage representation and previously decoded words. We first initialize the hidden state of the decoder GRU by passing the last backward encoder hidden state $\overleftarrow{h}_1$ to a linear layer:
\begin{equation}
\label{eq:decinit}
s_0 = tanh(\textbf{W}_0\overleftarrow{h}_1 + b).
\end{equation}

For each decoding time step $t$, the GRU reads the embedding of the previous word $w_{t-1}$, previous attentional context vector $c_{t-1}$, and its previous hidden state $s_{t-1}$ to calculate its current hidden state:
\begin{align}
\label{eq:decoder-update}
s_t = \mathbf{GRU}([w_{t-1}; c_{t-1}], s_{t-1}).
\end{align}

The context vector $c_t$ for time step $t$ is generated through the concatenated attention mechanism \cite{luong2015effective}. Attention mechanism calculates a semantic match between encoder hidden states and the decoder hidden state. The attention weights indicate how the model spreads out the amount it cares about different encoder hidden states during decoding. At time step $t$, the attention weights and the context vector are calculated as:
\begin{align}
\label{eq:attention}
e_{t,i} &= v^\intercal tanh(\mathbf{W}_s s_{t} + \mathbf{W}_h h_i),\\
\alpha_{t,i} &= \frac{\text{exp}(e_{t,i})}{\sum_{j=1}^{|P|}\text{exp}(e_{t,j})},\\
c_t &= \sum_{i=1}^{|P|}\alpha_{t,i} h_i.
\end{align}

Combining the previous word embedding $w_{t-1}$, the current decoder state $s_t$ and the current context vector $c_t$, we can calculate a readout state $r_t$ by an MLP maxout layer with dropouts \cite{goodfellow2013maxout}. Then the readout state is passed to a linear layer and a softmax layer to predict the probabilities of the next word over the decoder vocabulary:
\begin{align}
\label{eq:generator}
r_t &= \mathbf{W}_{rw} w_{t-1} + \mathbf{W}_{rc} c_t + \mathbf{W}_{rs} s_t\\
m_t &= [\text{max}\{r_{t,2j-1}, r_{t, 2j}\}]_{j=1,...,d}^\intercal\\
p(y_t|& y1,\cdots, y_{t-1}) = \text{softmax}(\mathbf{W}_o m_t),
\end{align}
where $r_t$ is a $2$-D vector.

The above module generates question words from a given vocabulary. Another important method to generate words is copying from source text. Copy or point mechanism \cite{gulcehre2016pointing} was introduced into sequence-to-sequence models to allow the copying of unknown words from the input text, which solves the out-of-vocabulary (OOV) problem.
When decoding at time step $t$, the probability of copying is given by:
\begin{align}
g_c = \sigma(\mathbf{W}_{cs} s_t + \mathbf{W}_{cc} c_t + b),
\end{align}
where $\sigma$ is the Sigmoid function, and $g_c$ is the probability of copying. For the copy probability of each input word, we reuse the attention weights given by Equation (8).

Copying mechanism has also been used in question generation \cite{zhou2017neural,song2018leveraging,hu2018aspect}, however, mainly to solve the OOV issue.
Here we leverage the copying mechanism to enable the copying of potential clue words, instead of being limited to OOV words, from input.
Different from existing methods, we take a more aggressive approach to train the copying mechanism via multitask learning based on different labels. That is, when preparing the training dataset, we explicitly label a word in a target question as a word copied from the source passage text if it satisfies all the following criteria:
\begin{description}
\item{i)} it appears in both source text and target question; 
\item{ii)} it is not a stop word; 
\item{iii)} its frequency rank in the vocabulary is lower than a threshold $r_h$.
\end{description} 
The remaining words in the question are considered as being generated from the vocabulary.
Such a binary label (copy or not copy), together with which input word the question word is copied from, as well as the target question, are fed into different parts of the decoder as labels for multi-task model training. This is to intentionally encourage the copying of potential clue words into the target question rather than generating them from a vocabulary.
  
The intuition behind such an aggressive copying mechanism can be understood by checking the frequency distributions of both generated words and copied words (as defined above) in the training dataset of SQuAD. Fig.~\ref{fig:rank-hist} shows the frequency distributions of all question words, and then of generated question words and copied question words. The mean and median rank of generated words are 2389 and 1032, respectively. While for copied words, they are 3119 and 1442. Comparing Fig.~\ref{fig:not_overlap_ranks} with Fig.~\ref{fig:non_stopword_overlap_ranks}, we can see that question words generated from the vocabulary tends to be clustered toward high ranked (or frequent) words. On the other hand, the fraction of low ranked (or infrequent) words in copied words are much greater than that in generated words. This means the generated words are mostly from frequent words, while the majority of low-frequency words in the long tail are copied from the input, rather than generated. 

This phenomenon matches our intuition: when people ask a question about a given passage, the generated words tend to be commonly used words, while the repeated text chunks from source passage may contain relatively rare words such as names and dates. Based on this observation, we further propose to reduce the vocabulary at the decoder for target question generation to be top $N$ frequently generated words, where $N$ is a predefined threshold that varies according to datasets.

\subsection{A GCN-Based Clue Word Predictor}
\label{subsec:cluepredict}

Given a passage and some answer positions, our clue word predictor aims to identify the clue words in the passage that can help to ask a question and are also potential candidates for copying, by understanding the semantic context of the input passage.
Again, in the training dataset, the non-stop words that are shared by both an input passage and an output question are aggressively labeled as \emph{clue words}.

\begin{figure*}[t]
                        \centering
                        \subfigure[Distribution of Syntactic Dependency Types between Copied Words and the Answer]{
                \includegraphics[width=2.2in]{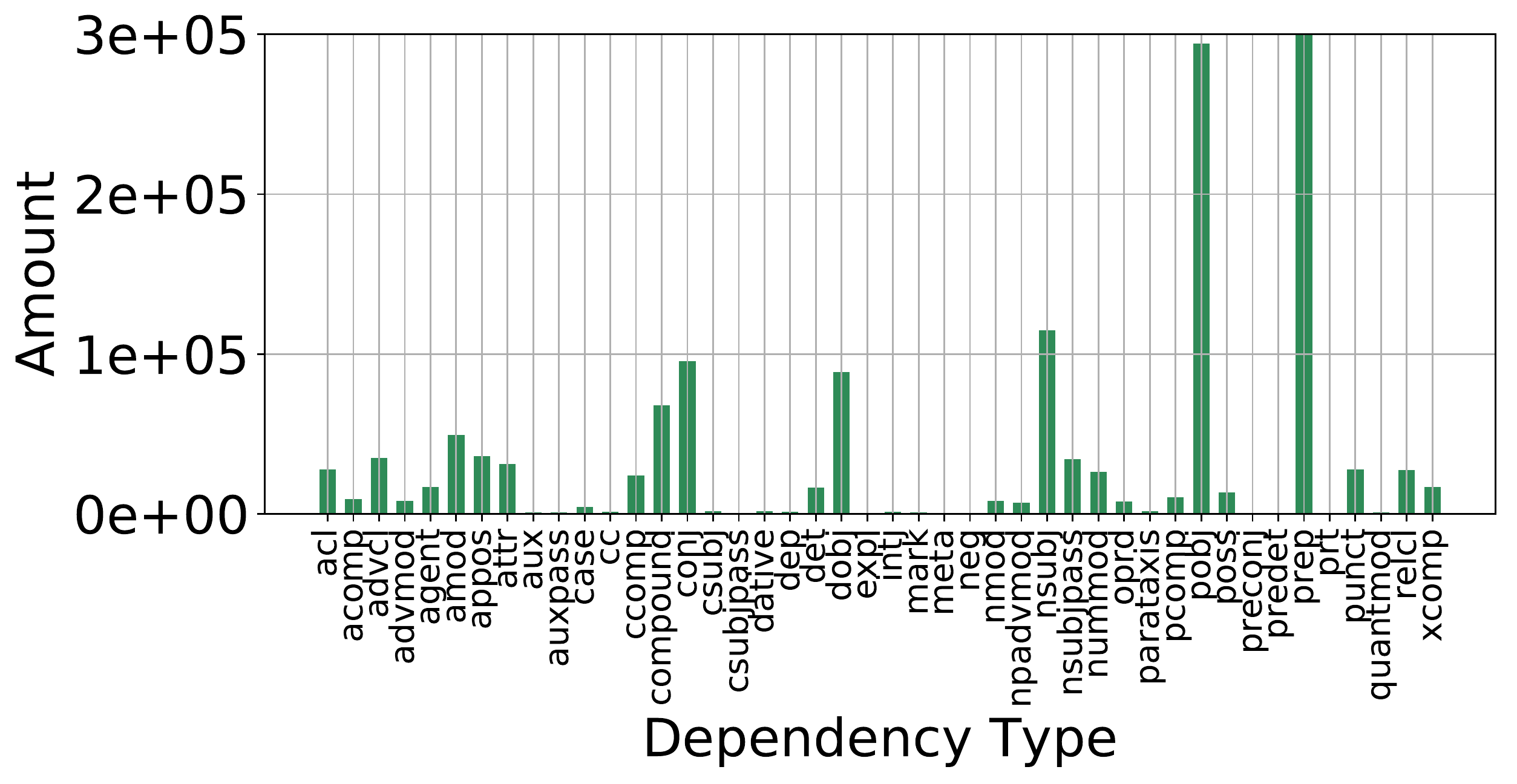}
                                \label{fig:bar_dep_type}
                        }
                \hspace{-1mm}
                        \subfigure[Distribution of Syntactic Dependency Distances between Copied Words and  theAnswer]{
                \includegraphics[width=2.2in]{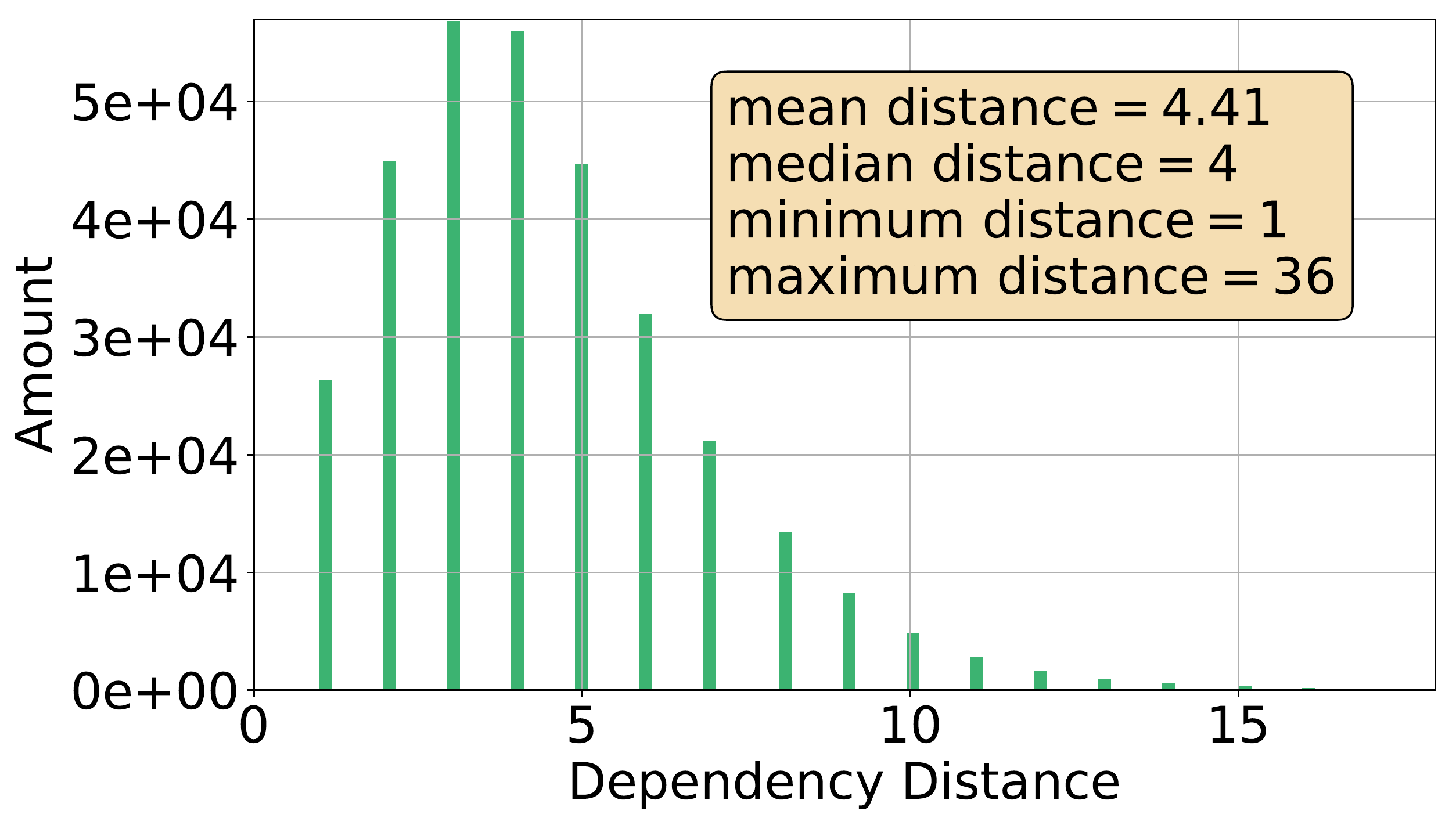}
                                \label{fig:dep_dist}
                        }
                        \hspace{-1mm}
                        \subfigure[Distribution of Sequential Word Distances between Copied Words and the Answer]{
                \includegraphics[width=2.2in]{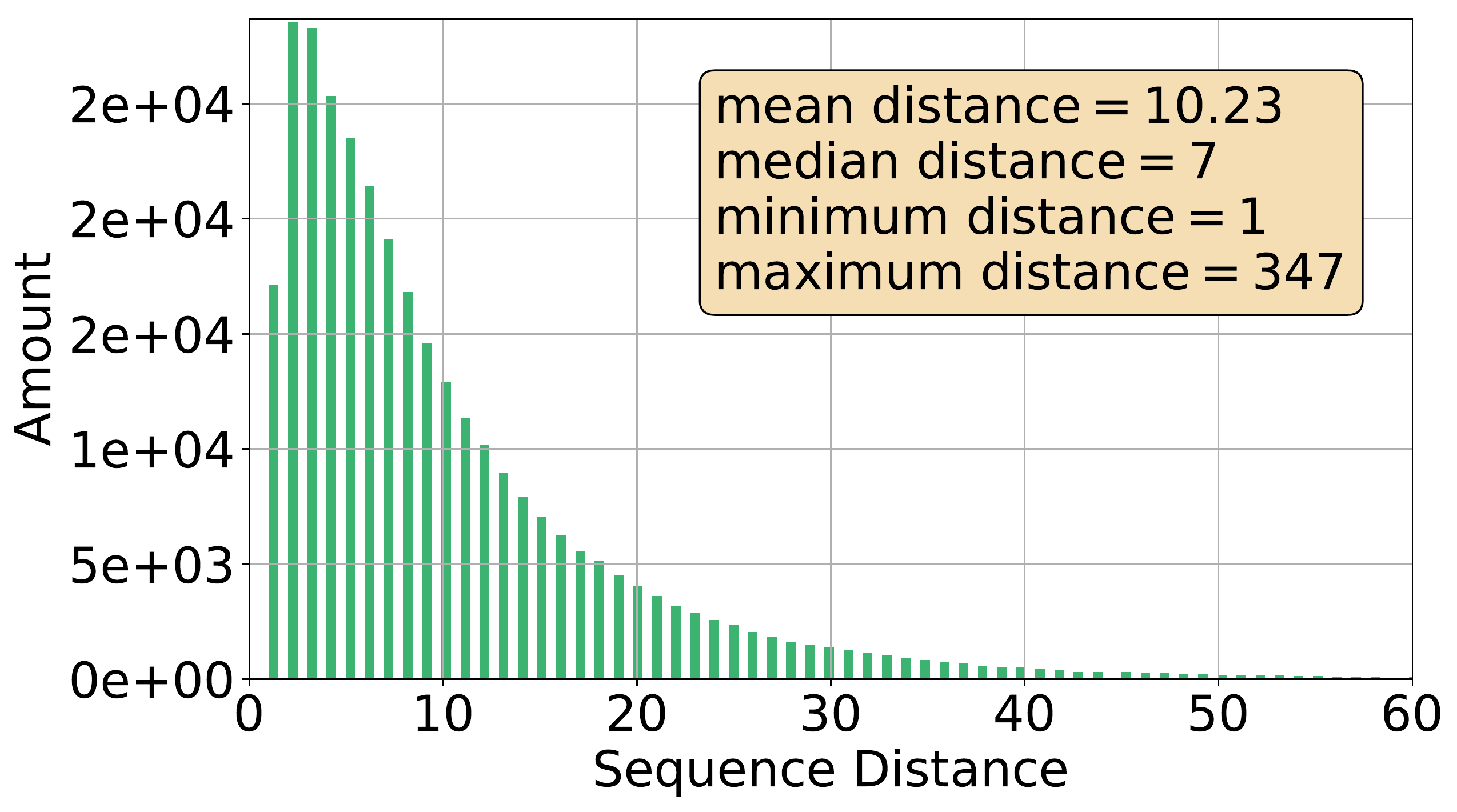}
                                \label{fig:word_dist}
                        }
                \vspace{-0mm}
                \caption{Comparing the distributions of syntactic dependency distances and sequential word distances between copied words and the answer in each training sample.} 
                \label{fig:distance}
\vspace{-2mm}
\end{figure*}

We note that clue words are, in fact, more closely connected to the answer chunk in the form of syntactic dependency trees than in word sequences.
Fig.~\ref{fig:distance} shows our observation on the SQuAD dataset. For each training example, we get the nonstop words that appear in both the input passage and the output question. For each clue word in the training set, we find its shortest undirected path to the answer chunk based on the dependency parsing tree of the passage. For each jump on the shortest path, we record the dependency type. We also calculate the distance between each clue word and the answer in terms of the number of words between them. As shown in Fig.~\ref{fig:bar_dep_type}, \textit{prep}, \textit{pobj}, and \textit{nsubj} appear frequently on these shortest paths. Comparing Fig.~\ref{fig:dep_dist} with Fig.~\ref{fig:word_dist}, we can see that the distances in terms of dependency trees are much smaller than those in terms of sequential word orders. The average and median distances on the dependency trees are $4.41$ and $4$, while those values  are $10.23$ and $7$ for sequential word distances.

In order to predict the positions of clue words based on their dependencies on the answer chunk (without knowing the question which is yet to be generated), we use a Graph Convolutional Network (GCN) to convolve over the word features on the dependency tree of each passage, as shown in   Fig.~\ref{fig:system}. The predictor consists of four layers: 

\textit{Embedding layer.} This layer shares the same features with the passage encoder, except that it does not include the clue indicators. Therefore, each word is represented by its word embedding, lexical features, binary features, the word frequency feature, and an answer position indicator.

\textit{Syntactic dependency parsing layer.} We obtain the syntactic dependency parsing tree of each passage by spaCy \cite{spacy}, where the dependency edges between words are directed. In our model, we use the syntactic structure to represent the structure of passage words.

\textit{Encoding layer.} The objective of the encoding layer is to encode the context information into each word based on the dependency tree. We utilize a multi-layered GCN to incorporate the information of neighboring word features into each vertex, which is a word.
After $L$ GCN layers, the hidden vector representation of each word will incorporate the information of its neighboring words that are no more than $L$ hops away in the dependency tree.

\textit{Output layer.} After obtaining the context-aware representation of each word in the passage, we calculate the probability of each word being a clue word. A linear layer is utilized to calculate the unnormalized probabilities. We subsequently sample the binary clue indicator for each word through a Gumbel-Softmax layer, given the unnormalized probabilities. A sampled value of $1$ indicates that the word is predicated as a clue word. 

\subsubsection{GCN Operations on Dependency Trees}

We now introduce the operations performed in each GCN layer~\cite{kipf2016semi,zhang2018graph}.
GCNs generalize the CNN from low-dimensional regular grids to high-dimensional irregular graph domains.
In general, the input to a GCN is a graph $G = (\mathcal{V}, E)$ with $N$ vertices $v_i \in \mathcal{V}$, and edges $e_{ij} = (v_i, v_j) \in E$. The edges can be weighted with weights $w_{ij}$, or unweighted. The input also contains a vertex feature matrix denoted by $X = \{\mathbf x_i\}_{i=1}^{N}$, where $\mathbf x_i$ is the \emph{feature vector} of vertex $v_i$. 

Since a dependency tree is also a graph, we can perform the graph convolution operations on dependency trees by representing each tree into its corresponding adjacency matrix form.
Now let us briefly introduce the GCN propagation layers used in our model \cite{zhang2018graph}.
The weighted adjacency matrix of the graph is denoted as $A \in \mathbb{R}^{N\times N}$ where $A_{ij} = w_{ij}$. For unweighted graphs, the weights are either 1 or 0. In an $L$-layer GCN, let $h_i^{(l-1)}$ denotes the input vector and $h_i^{(l)}$ denotes the output vector of node $i$ at the $l$-th layer. We will utilize a multi-layer GCN with the following layer-wise propagation rule~\cite{zhang2018graph}:
\begin{equation}
\label{eq:gcn}
h_i^{(l)} = \sigma(\sum_{j=1}^N \tilde{\mathbf{A}}_{ij}  \mathbf{W}^{(l)} h_j^{(l-1)} / d_i + b^{(l)}) ,
\end{equation}
where $\sigma$ is a nonlinear function (e.g., ReLU), and
$ W^{(l)}$ is a linear transformation.
$\tilde{\mathbf{A}} = \mathbf{A} + \mathbf{I}_N$ where $\mathbf{I}_N$ is an $n\times n$ identity matrix. $d_{i} = \sum_{j=1}^N \tilde{\mathbf{A}_{ij}}$ is the degree of node (or word in our case) $i$ in the graph (or dependency tree).

In our experiments, we treat the dependency trees as undirected, i.e., $\forall i,j, \mathbf{A}_{ij} = \mathbf{A}_{ji}$. Besides, as we already included the dependency type information in the embedding vectors of each word, we do not need to incorporate the edge type information in the adjacency matrix.

Stacking this operation by $L$ layers gives us a deep GCN network. The input to the nodes in the first layer of the GCN are the feature vectors of words in the passage. After $L$ layers of transformations, we can obtain a context-aware representation of each word. We then feed them into a linear layer to get the unnormalized probability of each word being a clue word. After that, the unnormalized probabilities are fed to a Straight-Through (ST) Gumbel-Softmax estimator to sample an $N$-dimensional binary vector indicating whether each of the $N$ words is a clue word or not.

\subsubsection{Gumbel-Softmax}

Gumbel-Softmax \cite{jang2016categorical} is a method
of sampling discrete random variables in neural networks. It approximates one-hot vectors sampled from a categorical distribution by making them continuous, therefore the gradients of model parameters can be calculated using the reparameterization trick and the standard backpropagation. 
Gumbel-Softmax distribution is motivated by Gumbel-Max trick \cite{maddison2014sampling}, an algorithm for sampling from a categorical distribution.
Let $(p_1, ..., p_k)$ denotes a $k$-dimensional categorical distribution where the probability $p_i$ of class $i$ is defined as:
\begin{align}
p_i = \frac{\exp(\log(\pi_i))}{\sum_{j=1}^k\exp(\log(\pi_j))},
\end{align}
where $\pi_i$ is the unnormalized log probability of class $i$. We can easily draw a one-hot sample $\mathbf{z} = (z_1, \cdots, z_k) \in \mathbb{R}^k$ from the distribution by the following equations:
\begin{align}
z_i &= \begin{cases}
    1,\  i = \argmax_j(log(\pi_j) + g_j)\\
    0,\  \text{otherwise}
  \end{cases}\\
  g_i  &= -\log(-\log(u_i)),\\
  u_i  &\sim \text{Uniform}(0, 1)
\end{align}
where $g_i$ is Gumbel noise used to perturb each $\log(\pi_i)$. In this way, taking $\argmax$ is equivalent to drawing a sample using probabilities $(p_1, \cdots, p_k)$.

Gumbel-Softmax distribution replaces the $\argmax$ function by differentiable softmax function. Therefore, a sample $\mathbf{y} = (y_1, \cdots, y_k)$ drawn from Gumbel-Softmax distribution is given by:
\begin{align}
y_i = \frac{\exp((\log(\pi_i) + g_i)/ \tau)}{\sum_{j=1}^k \exp((\log(\pi_j) + g_j)/ \tau)},
\end{align}
where $\tau$ is a temperature parameter. The Gumbel-Softmax distribution resembles the one-hot sample when $\tau$ diminishes to zero.

Straight-Through (ST) Gumbel-Softmax estimator \cite{jang2016categorical} is a discrete version of the continuous Gumbel-Softmax estimator. It takes different paths in the forward and backward propagation.
In the forward pass, it discretizes a continuous probability vector $\mathbf{y}$ sampled from the Gumbel-Softmax distribution into a one-hot vector $\mathbf{y}^{ST} = (y_1^{ST}, \cdots, y_k^{ST})$ by:
\begin{align}
y_i^{ST} &= \begin{cases}
    1,\  i = \argmax_j y_j,\\
    0,\  \text{otherwise}.
  \end{cases}
\end{align}
In the backward pass, it uses the continuous $\mathbf{y}$,
so that the error signal can still backpropagate.

Using the ST Gumbel-Softmax estimator, our model is able to sample a binary clue indicator vector for an input passage. Then the clue indicator vector is fed into the passage encoder for question generations, as shown in Fig.~\ref{fig:system}.

\section{Evaluation}
\label{sec:simu}

In this section, we evaluate the performance of our proposed models on the SQuAD dataset and the NewsQA dataset, and compare them with state-of-the-art question generation models.

\subsection{Datasets, Metrics and Baselines}

\begin{table}[tb]
\small
  \resizebox{8cm}{!}
  {
    \begin{threeparttable}
      \caption{Description of evaluation datasets.}
      \label{tab:datasets}
      \begin{tabular}{lllllll}
        \toprule
        Dataset & Train & Dev & Test & P-Length$^*$ & Q-Length$^*$ & A-Length$^*$ \\
        \midrule
        SQuAD &  $86,635$ & $8,965$ & $8,964$ & $32.72$ & $11.31$ & $3.19$\\
        NewsQA &  $77,538$ & $4,341$ & $4,383$  & $28.14$ & $7.82$ & $4.60$\\
        \bottomrule
      \end{tabular}
      \begin{tablenotes}
        \small
        \item
        $^*$ P-Length: average number of tokens of passages.\\
        $^*$ Q-Length: average number of tokens of questions.\\
        $^*$ A-Length: average number of tokens of answers.
      \end{tablenotes}
      \vspace{-4mm}
    \end{threeparttable}
  }
\end{table}

The SQuAD dataset is a reading comprehension dataset, consisting of questions posed by crowd-workers on a set of Wikipedia articles, where the answer to every question is a segment of text from the corresponding reading passage. SQuAD 1.1 is used in our experiment containing 536 Wikipedia articles and more than 100K question-answer pairs. When processing a sample from dataset, instead of using the entire document, we take the sentence that contains the answer as the input. Since the test set is not publicly available, we use the data split proposed by \cite{zhou2017neural} where the original dev set is randomly split into a dev test and a test set of equal size. 

In the NewsQA dataset, there are 120K questions and their corresponding answers as well as the documents that are CNN news articles. Questions are written by questioners in natural language with only the headlines and highlights of the articles available to them. With the information of the questions and the full articles, answerers select related sub-spans from the passages of the source text and mark them as answers. Multiple answers may be provided to a same question by different answerers and they are ranked by validators based on the quality of the answers. In our experiment, we picked a subset of NewsQA where answers are top-ranked and are composed of a contiguous sequence of words within the input sentence of the document. 

Table \ref{tab:datasets} shows the number of samples in each set and the average number of tokens of the input sentences, questions, and answers listed in columns P-Length, Q-Length, and A-Length respectively. \\

We report the evaluation results with following metrics.

\begin{itemize}
	\item \textbf{BLEU} \cite{papineni2002bleu}. 
	BLEU measures precision by how much the words in prediction sentences appear in reference sentences at the corpus level. BLEU-1, BLEU-2, BLEU-3, and BLEU-4, use 1-gram to 4-gram for calculation, respectively.
	\item \textbf{ROUGE-L} \cite{lin2004rouge}.
	ROUGE-L measures recall by how much the words in reference sentences appear in prediction sentences using Longest Common Subsequence (LCS) based statistics.
	\item \textbf{METEOR} \cite{denkowski2014meteor}. 
	METEOR is based on the harmonic mean of unigram precision and recall, with recall weighted higher than precision.
\end{itemize}

In the experiments, we have eight baseline models for comparison. Results reported on PCFG-Trans, MPQG, and NQG++ are from experiments we conducted using published code on GitHub. For other baseline models, we directly copy the reported performance given in their papers. We report all the results on the SQuAD dataset, and for the NewsQA dataset, we can only report the baselines with open source code available.

\begin{itemize}
	\item \textbf{PCFG-Trans} \cite{heilman2011automatic} is a rule-based system that generates a question based on a given answer word span.
	\item \textbf{MPQG} \cite{song2018leveraging} proposed a Seq2Seq model that matches the answer with the passage before generating the question.
	\item \textbf{SeqCopyNet} \cite{zhou2018sequential} proposed a method to improve the copying mechanism in Seq2Seq models by copying not only a single word but a sequence of words from the input sentence.
	\item \textbf{seq2seq+z+c+GAN} \cite{yao2018teaching} proposed a model employed in GAN framework using the latent variable to capture the diversity and learning disentangled representation using the observed variable.
	\item \textbf{NQG++} \cite{zhou2017neural} proposed a Seq2Seq model with a feature-rich encoder to encode answer position, POS and NER tag information.
	\item \textbf{Answer-focused Position-aware model} \cite{sun2018answer} incorporates the answer embedding to help generate an interrogative word matching the answer type. And it models the relative distance between the context words and the answer for the model to be aware of the position of the context words when generating a question.
	\item \textbf{s2sa-at-mp-gsa} \cite{zhao2018paragraph} proposed a model which contains a gated attention encoder and a maxout pointer decoder to address the challenges of processing long text inputs. This model has a paragraph-level version and a sentence-level version. For the purpose of fair comparison, we report the results of the sentence-level model to match with our settings.
	\item \textbf{ASs2s} \cite{kim2018improving} proposed an answer-separated Seq2Seq to identify which interrogative word should be used by replacing the target answer in the original passage with a special token.
\end{itemize}

For our models, we evaluate the following versions:
\begin{itemize}
    \item \textbf{CGC-QG (no feature-rich embedding)}. We name our model as Clue Guided Copy for Question Generation (CGC-QG). In this variant, we only keep the embedding of words, answer position indicators, and clue indicators for each token, and remove the embedding vectors of other features.
    \item \textbf{CGC-QG (no target reduction)}. This model variant does not contain target vocabulary reduction operation.
    \item \textbf{CGC-QG (no clue prediction)}. The clue predictor and clue embedding are removed in model variant.
    \item \textbf{CGC-QG}. This is the complete version of our proposed model.
\end{itemize}

\subsection{Experiment Settings}

We implement our models in PyTorch 0.4.1 \cite{paszke2017pytorch} and train the model with a single Tesla P40. We utilize spaCy \cite{spacy} to perform dependency parsing and extract lexical features for tokens. As to the vocabulary, we collect it from the training dataset and keep the top 20K most frequent words for both datasets.

We set the threshold $r_h = 100$ and $r_l = 2000$. For target vocabulary reduction, we set $N = 2000$. The embedding dimension of word vector is set to be $300$ and initialized by GloVe. The word vectors of words that are not contained in GloVe are initialized randomly. The word frequency features are embedded to 32-dimensional vectors, and other features and indicators are embedded to 16-dimensional vectors. All embedding vectors are trainable in the model. We use a single layer BiGRU with hidden size 512 for the encoder, and a single layer undirected GRU with hidden size 512 for the decoder. 
The dropout rate $p = 0.1$ is applied to the encoder, decoder, and the attention module.

During training, we optimize the Cross-Entropy loss function for clue prediction, question generation, and question copying, and perform gradient descent by the Adam \cite{kingma2014adam} optimizer with an
initial learning rate $l_r =0.001$, two momentum parameters are $\beta_1=0.8$ and $\beta_2=0.999$ respectively, and $\epsilon=10^{-8}$. The mini-batch size for each update is set to 32 and model is trained for up to 10 epochs (as we found that usually the models derive the best performance after $6$ or $7$ epochs). We apply gradient clipping with range $[-5, 5]$ for Adam. Besides, exponential moving average is applied on all trainable variables with a decay rate $0.9999$.
When testing, beam search is conducted with beam width $20$. The decoding process stops when a token \textit{<EOS>} that represents end of sentence is generated.

\subsection{Main Results}

Table.~\ref{tab:squad-dataset} and Table.~\ref{tab:newsqa-dataset} compare the performance of our model with existing question generation models on SQuAD and NewsQA in terms of different evaluation metrics.
For the SQuAD dataset, we compare our model with all the baseline methods we have listed. As to the NewsQA dataset, since only a part of the baseline methods made their code public, we compare our model with approaches that have open source code.
We can see that our model achieves the best performance on both datasets and significantly outperforms state-of-the-art algorithms.
On the SQuAD dataset, given an input sentence and an answer, the BLEU-4, ROUGE-L, and METEOR of our result are $17.55$, $44.53$, and $21.24$ respectively, while corresponding previous state-of-the-art results are $16.17$, $44.24$, and $19.67$ from different approaches. Similarly, our method also gives a significantly better performance on NewsQA compared with the baselines in Table \ref{tab:newsqa-dataset}.

The reason is that we combine different strategies in our model to make it learn when to generate or copy a word, and what to generate or copy. First, our model learns to predict clue words through a GCN-based clue predictor. Second, our encoder incorporates a variety of embeddings of different features and clue indicators. Combining with the masking strategy, our model can better discover the relationship between input patterns and output patterns. Third, the reduced target vocabulary also helps our model to better capture when to copy or generate, and the generator is easier to train with a reduced vocabulary size. And most importantly,
our new criteria of marking a question word as copied word (as described in Sec.\ref{subsec:cluepredict}) helps the model to make better decisions on which path to go, i.e., to copy or to generate, during question generation.
By incorporating part of these new strategies and modules into our model, we can achieve performance better than state-of-the-art models on SQuAD and NewsQA. With all these designs implemented, our model gives the best performance on both datasets.

There is a significant gap between the performance on SQuAD and on NewsQA due to the different characteristics of the datasets. The average answer length of NewsQA is 44.2\% larger than it of SQuAD according to the statistics shown in Table \ref{tab:datasets}. Long answers usually hold more information and are more difficult to generate questions. Furthermore, reference questions in NewQA tend to have less strict grammars and more diverse phrasings. To give a typical example, ``Iran criticizes who?'' is a reference question in NewsQA which does not start with an interrogative word but ends with one. These characteristics make the performance on NewsQA not as good as on SQuAD. However, our approach is still significantly better than the compared approaches on NewsQA dataset. It demonstrates that copy from the input is a general phenomenon across different datasets. Our model better captures what copied words are and what generated words are in a question based on our new criteria of labeling a question word as copied word or not.

\begin{table*}[tb]
\small
  \caption{Evaluation results of different models on SQuAD dataset.}
  \label{tab:squad-dataset}
  \begin{threeparttable}
  \begin{tabular}{c|cccccc}
    \toprule
    \textbf{Model} & \textbf{BLEU-1} & \textbf{BLEU-2} & \textbf{BLEU-3} & \textbf{BLEU-4} & \textbf{ROUGE-L} & \textbf{METEOR} \\
    \midrule
    PCFG-Trans$^*$ & $28.77$ & $17.81$ & $12.64$ & $9.47$ & $31.68$ & $18.97$ \\
    SeqCopyNet & $-$ & $-$ & $-$ & $13.02$ & $44.00$ & $-$ \\
    seq2seq+z+c+GAN & $44.42$ & $26.03$ & $17.60$ & $13.36$ & $40.42$ & $17.70$ \\
    NQG++$^*$ & $42.36$ & $26.33$ & $18.46$ & $13.51$ & $41.60$ & $18.18$ \\
    MPQG & $-$ & $-$ & $-$ & $13.91$ & $-$ & $-$ \\
    Answer-focused Position-aware model & $43.02$ & $28.14$ & $20.51$ & $15.64$ & $-$ & $-$ \\
    s2sa-at-mp-gsa & $44.51$ & $29.07$ & $21.06$ & $15.82$ & $44.24$ & $19.67$ \\
    ASs2s & $-$ & $-$ & $-$ & $16.17$ & $-$ & $-$ \\
    \midrule
    CGC-QG (no feature-rich embedding) & $45.50$ & $29.63$ & $21.58$ & $16.38$ & $43.11$ & $20.52$ \\
    CGC-QG (no target reduction) & $45.80$ & $30.27$ & $22.29$ & $17.05$ & $44.09$ & $20.79$ \\
    CGC-QG (no clue prediction) & $45.58$ & $30.07$ & $22.08$ & $16.80$ & $44.51$ & $20.80$ \\
    CGC-QG & $\mathbf{46.58}$ & $\mathbf{30.90}$ & $\mathbf{22.82}$ & $\mathbf{17.55}$ & $\mathbf{44.53}$ & $\mathbf{21.24}$\\
    \bottomrule
  \end{tabular}
  \begin{tablenotes}
    \small
    \item Experiments are conducted on baselines followed by a ``$^*$'' using released source code. Results of other baselines are copied from their papers where unreported metrics are marked ``$-$''.
  \end{tablenotes}
  \vspace{-4mm}
  \end{threeparttable}
\end{table*}

 \begin{table*}[tb]
 \small
   \caption{Evaluation results of different models on NewsQA dataset.}
   \label{tab:newsqa-dataset}
   \begin{threeparttable}
   \begin{tabular}{c|cccccc}
     \toprule
     \textbf{Model} & \textbf{BLEU-1} & \textbf{BLEU-2} & \textbf{BLEU-3} & \textbf{BLEU-4} & \textbf{ROUGE-L} & \textbf{METEOR} \\
     \midrule
     PCFG-Trans$^*$ & $16.90$ & $7.94$ & $4.72$ & $3.08$ & $23.78$ & $13.74$ \\
     MPQG$^*$ & $35.70$ & $17.16$ & $9.64$ & $5.65$ & $39.85$ & $14.13$ \\
     NQG++$^*$ & $40.33$ & $22.47$ & $14.83$ & $9.94$ & $42.25$ & $16.72$ \\
     \midrule
     CGC-QG (no feature-rich embedding) & $39.35$ & $22.10$ & $14.36$ & $9.99$ & $42.00$ & $16.60$ \\
     CGC-QG (no target reduction) & $40.00$ & $22.84$ & $15.01$ & $10.52$ & $42.33$ & $16.89$ \\
     CGC-QG (no clue prediction) & $39.85$ & $22.82$ & $14.96$ & $10.45$ & $43.16$ & $17.11$ \\
     CGC-QG & $\mathbf{40.45}$ & $\mathbf{23.52}$ & $\mathbf{15.68}$ & $\mathbf{11.06}$ & $\mathbf{43.16}$ & $\mathbf{17.43}$\\
     \bottomrule
   \end{tabular}
   \begin{tablenotes}
    \small
    \item Experiments are conducted on baselines followed by a ``$^*$'' using released source code.
   \end{tablenotes}
   \vspace{-4mm}
   \end{threeparttable}
 \end{table*}

\subsection{Analysis}

We evaluate the impact of different modules in our model by ablation tests. Table \ref{tab:squad-dataset} and Table \ref{tab:newsqa-dataset} list the performance of our model variants with different sub-components removed.

When we remove the extra feature embeddings in our model, i.e., the embeddings of POS, NER, Dependency Types, word frequency levels (low-frequent, median-frequent, high-frequent), and binary features (whether it is lowercase, digit), the performance drops significantly. This is because the tags and feature embeddings represent each token in different aspects. The number of different tags is much smaller than the number of different words. Therefore, the patterns which can be learned from these tags and features are more obvious than what we can learn from word embeddings. Even though a well-trained word vector may contain the information of other features such as POS or NER, explicitly concatenating these feature embedding vectors helps the model to capture the patterns to ask a question more easily.

Removing the operation of target vocabulary reduction also hurts the performance of our model. As we discussed earlier, the non-overlap question words (or generated words) are mostly covered by the high frequency words. Reducing the target vocabulary size helps our model to better learn the probabilities of generating these words. On the other hand, it also encourages the model to better capture what they can copy from input text.

Finally, without the clue prediction module, the performance also drops on both datasets. This is because when given an answer span in a passage, asking a question about it is still a one-to-many mapping problem. Our clue prediction module learns how people select the related clue words to further reduce the uncertainty of how to ask a question by learning from a large training dataset. With predicted clue indicators incorporated into the encoder of generator, our model can fit the way how people ask questions in the dataset.

\section{Related Work}
\label{sec:related}

In this section, we review related works on question generation and graph convolutional networks.

Existing approaches for question generation can be mainly classified into two classes: heuristic rule-based approaches and neural network-based approaches.
The rule-based approaches rely on well-designed rules or templates manually created by human to transform a piece of given text to questions \cite{heilman2010good,heilman2011automatic,chali2015towards}.
However, they require creating rules and templates by experts which is extremely expensive. Also, rules and templates have a lack of diversity and are hard to generalize to different domains.

Compared with rule-based approaches, neural network-based models are trained end-to-end and do not rely on hand-crafted rules or templates. Most of the neural question generation models consider the task as Seq2Seq and take advantage of the encoder-decoder framework with attention mechanism. \cite{serban2016generating} utilizes an encoder-decoder framework with attention mechanism to generate factoid questions from FreeBase. \cite{du2017learning} generates questions from SQuAD passages based on a sequence-to-sequence model with attention mechanism.

However, given an input sentence, generating questions is a one-to-many mapping, as we can ask different questions from different aspects. Purely relying on Seq2Seq model may not be able to learn such a one-to-many mapping \cite{gao2018difficulty}. To resolve this issue, recent works assume the aspect is known when generating a question \cite{zhou2017neural,yuan2017machine,subramanian2017neural,kim2018improving,song2018leveraging} or can be detected by a third-party pipeline \cite{du2018harvesting}. \cite{zhou2017neural} enriches the sequence-to-sequence model with answer position indicator to indicate if the current word is an answer word or not, and further incorporates copy mechanism to copy words from the context when generating a question. \cite{song2018leveraging,hu2018aspect} fuse the answer information into input sentence first, and apply a Seq2Seq model with attention and copy mechanism to generate answer-aware questions. \cite{gao2018difficulty} takes question difficulty into account by a difficulty estimator, and generate questions on different difficulty levels. \cite{sun2018answer} enriches the model with both answer embedding and the relative distance between the context words and the answer. \cite{tang2017question,tang2018learning} model question answering and question generation as dual tasks. They found that jointly training the two tasks helped to generate better questions.

We argue that even with an answer indicator, the problem of question generation is still a one-to-many mapping. To solve this problem, we enrich the model with a ``clue word'' predictor, where a clue word means a word that is related to the aspect of the targeting output question and usually copied to the question. We represent an input sentence by its syntactic dependency tree and represent each word using feature-rich embeddings, and let the model learn to predict whether each word in context can be a clue word or not. Combining the clue prediction module with a Seq2Seq model with attention and copy mechanism, our model learns when to generate a word from a target vocabulary and when to copy a word from the input context.

Graph Convolutional Networks generalize Convolutional Neural Networks to graph-structured data, and have been
developed and grown rapidly in scope and popularity in recent years \cite{kipf2016semi,defferrard2016convolutional,liu2018matching,marcheggiani2017encoding,battaglia2018relational}. Here we focus on the applications of GCNs on natural language. \cite{marcheggiani2017encoding} applies GCNs over syntactic dependency trees as sentence encoders, and produces latent feature representations of words in a sentence for semantic role labeling. \cite{liu2018matching} matches long document pairs using graph structures, and classify the relationships of two documents by GCN. \cite{zhang2018graph} proposes an extension of graph convolutional networks that is tailored for relation extraction. It pools information over dependency trees efficiently in parallel. In our paper, we apply GCN over the dependency tree of an input sentence, and predict the potential clue words in the sentence together with a Gumbel-Softmax estimator.


\section{Conclusion}
\label{sec:conclude}

In this paper, we demonstrate the effectiveness of teaching the model to make decisions during the question generation process on which words to generate and to copy.
We label the nonstop and overlap words between input passages and questions as copy targets and use such labels to train our model.
Besides, we further observe that the distribution of generated question words are mostly common words with relative high frequency.
Based on this observation, we reduce the vocabulary size for generating question words.
To help the model better capture how to ask a question and alleviate the issue of one-to-many mapping when asking a question, we propose a GCN-based clue prediction module to predict which part of words can be a clue word to ask a question given an answer. 
It utilizes the syntactic dependency tree representation of a passage to encode the information of each token in the passage, and sample a clue indicator for each token using a Straight-Through (ST) Gumbel-Softmax estimator. Our simulation results on the SQuAD dataset and NewsQA dataset show that our model outperforms a range of existing state-of-the-art approaches significantly.

\newpage
\bibliographystyle{ACM-Reference-Format}
\balance
\bibliography{main}

\end{document}